\documentclass{article}

\usepackage{arxiv}

\usepackage[utf8]{inputenc} 
\usepackage[T1]{fontenc}    
\usepackage{hyperref}       
\usepackage{url}            
\usepackage{booktabs}       
\usepackage{amsfonts}       
\usepackage{nicefrac}       
\usepackage{microtype}      
\usepackage{lipsum}
\usepackage{graphicx}
\usepackage{natbib}
\usepackage{amsmath}
\usepackage{float}
\usepackage[vlined]{algorithm2e}
\graphicspath{ {images/} }

\DontPrintSemicolon
\SetSideCommentLeft

\DeclareMathOperator*{\argmin}{arg\,min}
\DeclareMathOperator*{\sh}{shuffle}

\title{Supervised Topological Maps}

\author{
 Francesco Mannella \\
  Institute of Cognitive Sciences and Technologies\\
  National Research Council of Italy\\
  Via San Martino della Battaglia, 44 00185 Roma \\
  \texttt{francesco.mannella@istc.cnr.it} \\
}

\begin{document}
\maketitle
\begin{abstract}
Controlling the internal representation space of a neural network is a
desirable feature because it allows to generate new data in a supervised manner.
In this paper we will show how this can be achieved while building a
low-dimensional mapping of the input stream, by deriving a generalized
algorithm starting from Self Organizing Maps (SOMs).
SOMs are a kind of neural
network which can be trained with unsupervised learning to produce a
low-dimensional discretized mapping of the input space. They can be used for
the generation of new data through backward propagation of
interpolations made from the mapping grid.
Unfortunately the final topology of
the mapping space of a SOM is not known before learning, so interpolating
new data in a supervised way is not an easy task.
Here we will show a variation from the SOM algorithm consisting in constraining
the update of prototypes so that it is also a function of the distance of its
prototypes from extrinsically given targets in the mapping space. We will
demonstrate how such variants, that we will call Supervised Topological Maps
(STMs), allow for a supervised mapping where the position of internal
representations in the mapping space is determined by the experimenter.
Controlling the internal representation space in STMs reveals to be an easier
task than what is currently done using other algorithms such as variational or
adversarial autoencoders.
 \end{abstract}


\section{Introduction}
\label{sec:introduction}
Neural networks are a powerful tool because they can implement any kind of
function by mapping the input space into an output domain with different
dimensionality. This is done while extracting features from inputs so that the
activity of deeper layers (far from the input layer) is an abstract
representation of the more superficial ones. Nevertheless the way learning of
the internal weights is achieved, makes them as black boxes and internal
representations cannot be usefully deployed to build efficient and intuitive
classifiers. Solving the issue of the intelligibility of internal
representations requires two functionalities: first, the creation of internal
representations must follow a meaningful heuristic; second, the position of
representations in the layer must be known.
Self organizing maps \citep[SOMs --][]{Kohonen1982} are a way to implement the
first functionality.  SOMs are competitive neural networks that allow for a
low-dimensional discretized mapping of the input space. The training algorithm
for SOMs is strictly related to the vanilla k-means clustering algorithm
\citep{macqueen1967,bishop2006}.
We show here that both algorithm belong to a family of algorithms where each
individual is defined by a different implementation of a winner-takes-all (WTA)
operator which is a function of the distance between input patterns and
prototypes, but can also depend on other variables, intrinsic or extrinsic
w.r.t. the structure of the network.
In particular we can define a variant whose WTA operator is also a function of
the distance of prototypes from a desired point in the low-dimensional mapping
space. We call This kind of network Supervised Topological Maps (STM). We show
in this paper that controlling the internal representation space in STMs is
easier than what can be currently done using other algorithms such as
variational or adversarial autoencoders \citep{Kingma2013,Makhzani2015}. In the
rest of the paper we will first describe the k-mean and SOM algorithms focusing
on their strict relation (sections \ref{sec:k-means} and \ref{sec:soms}). Then
we define STMs by describing their specific WTA operator (section
\ref{sec:stms}). Finally we will show some examples of usage of STMs for the
supervised generation of data (section \ref{sec:results}) and we will discuss
their relationship with other algorithms and their limitations (section
\ref{sec:discussion}).

\section{k-means clustering}
\label{sec:k-means}
Vanilla k-means clustering divides the patterns of a dataset into $K$ clusters
where $K$ is a fixed parameter. Each iteration of the algorithm is composed of
two steps. In the first step input patterns are assigned to clusters based on
their distance to the cluster centroids. Each pattern is assigned to its closer
cluster. In the second step each $j^{th}$ centroid is substituted with the mean
of all patterns belonging to the $j^{th}$ cluster.

Formally, the k-means algorithm can be described as a minimization over an
energy function. Given a dataset $\mathbf{X} = \left[\mathbf{x}_1,\cdots,
\mathbf{x}_i, \cdots, \mathbf{x}_N\right]^T \in \mathcal{R}^{N \times M}$,  with
each input pattern $\mathbf{x}_i \in \mathcal{R}^M$,  we can define the first
step of the iteration as the application of WTA operator:
\begin{align}
  \label{eq:k-means-phi}
  r_i &= \underset{k}{\argmin} || \mathbf{x}_i -\mathbf{c}_k || \\
  \phi_{i,j} & = \left\{
  \begin{array}{ll}
    1 & \texttt{if}\ j = r_i\\
    0 & \texttt{otherwise}
  \end{array}
  \right.
\end{align}

and the definition the energy function based on the WTA operator:
\begin{equation}
  \label{eq:k-means-energy}
  L = \frac{1}{2}\sum\limits_{i=1}^{N}\sum\limits_{j=1}^{K}{\phi_{i, j}
  || \mathbf{x}_i - \mathbf{c}_j ||^2}
\end{equation}
In the second step we minimize the energy function w.r.t each centroid
$\mathbf{c}_j$:
\begin{align}
  \frac{\partial L}{\partial \mathbf{c}_j} &= -\sum\limits_{i=1}^{N}
  \sum\limits_{j=1}^{K}{\phi_{i, j} (\mathbf{x}_i - \mathbf{c}_j)} = 0 \\
  \mathbf{c}_j & = \frac{\sum\limits_{i=1}^{N}
  {\phi_{i, j}\mathbf{x}_i}}{\sum\limits_{i=1}^{N}\phi_{i, j}}
\end{align}
so that, as said before, each centroid $\mathbf{c}_j$ is updated as the mean of
all input patterns currently belonging to the $j^{th}$ cluster. The learning
process ends when the centroids do not change anymore between iterations and the
equilibrium is reached (see the full procedure in Algorithm \ref{alg:kmeans_batch}).

The update step can be also applied iteratively:

\begin{equation}
  \mathbf{c}_j = \mathbf{c}_j + \eta_i\phi_{i, j} (\mathbf{x}_i - \mathbf{c}_j)
\end{equation}

where $\eta_i$ is the learning rate parameter, which is typically made to
decrease monotonically as more data points are considered \citep{bishop2006}.
This iterative form highlights the strict relationship between k-means
clustering and SOMs (see Algorithm \ref{alg:kmeans_online} and Algorithm
\ref{alg:som_online}).

\begin{minipage}[c]{\textwidth}
\RestyleAlgo{ruled}
\begin{algorithm}[H]
\label{alg:kmeans_batch}
\KwIn{
    $\mathbf{X} \in \mathcal{R}^{N, M}$: input dataset\newline
    $\mathbf{C} \in \mathcal{R}^{K, M}$: centroids of the K clusters\newline
 }
\BlankLine
\While{$\sum\limits_{j=0}^K || \Delta \mathbf{c}_j || \ < \epsilon$}{
  \For{$i\gets1$ \KwTo N} {
    $r_{i} = \underset{k}{\argmin} || \mathbf{x}_i - \mathbf{c}_k ||$
    \tcp*[r]{Index of the winner.}
    \For{$j\gets1$ \KwTo $K$} {
      $\phi_{i, j}\gets
        \left\{
          \begin{array}{ll}
            1 & \texttt{if}\ j = r_i\\
            0 & \texttt{otherwise}
          \end{array}
          \right.$
      \tcp*[r]{Update operator.}
    }
  }

  \For{$j\gets1$ \KwTo $K$} {
    $\mathbf{c}_j \gets \frac{\sum\limits_{i=1}^N\phi_{i,j}\mathbf{x}_i}
    {\sum\limits_{i=1}^N\phi_{i,j}}$
    \tcp*[r]{Updating of clusters.}
  }
}
\caption{Vanilla k-means algorithm -- batch version}
\end{algorithm}
\end{minipage}

\begin{minipage}[c]{\textwidth} \RestyleAlgo{ruled}
\begin{algorithm}[H]
\label{alg:kmeans_online}
\KwIn{
  $\mathbf{X} \in \mathcal{R}^{N, M}$: input dataset\newline
  $\mathbf{C} \in \mathcal{R}^{K, M}$: centroids of the K clusters\newline
  $\eta_{init}$: initial learning rate \newline
  $\tau$: decay window \newline
  $T$: number of epochs
 }
\BlankLine
\For{$t\gets0$ \KwTo $T-1$}{
  $\eta\gets\eta_{init}e^{-\frac{t}{\tau}}$\;
  $\sh{\mathbf{X}}$\;
  \For{$i\gets1$ \KwTo N} {
    $r_{i} = \underset{k}{\argmin} || \mathbf{x}_i - \mathbf{c}_k ||$
    \tcp*[r]{Index of the winner.}
    \For{$j\gets1$ \KwTo $K$} {
      $\phi_{i, j}\gets
        \left\{
          \begin{array}{ll}
            1 & \texttt{if}\ j = r_i\\
            0 & \texttt{otherwise}
          \end{array}
          \right.$
      \tcp*[r]{Update operator.}
      $\mathbf{c}_j \gets \mathbf{c}_j + \eta\phi_{i,j}(\mathbf{x}_i
        - \mathbf{c}_j)$
      \tcp*[r]{Updating of clusters.}
    }
  }
}
\caption{Vanilla k-means algorithm -- online version}
\end{algorithm}
\end{minipage}

\section{Self Organizing Maps}
\label{sec:soms}
SOMs are competitive neural networks, where competition between the units of the
inner layer allows for the unsupervised emergence of a low-dimensional
discretized map.
One basic feature is that the inner layer has a predefined intrinsic topology,
for instance, units in the inner layer can disposed in a 1-, 2- or 3-dimensional
grid. After learning the weights of connections from the input layer to each
unit in the inner layer become a prototype (centroid) of a cluster within the
input dataset, in analogy with k-means cluster centroids. Similarly to k-means
clustering, the update of centroids depends on a WTA competition based on the
euclidean distance of input patterns from the centroids. Differently from
k-means clustering the update also depends on the euclidean distance of the
centroids from the winner centroid in the space of the inner layer. In
particular, a radial-basis function of the euclidean distance from the winner in
the inner layer space is used (will call it the neighboring function).
\begin{align}
  r_i &= \underset{k}{\argmin} || \mathbf{x}_i -\mathbf{w}_k || \\
  \label{eq:som_phi}
  \phi_{i, j} &= e^{-\frac{|| j - r_i ||^2}{2\sigma^2}}
\end{align}
where $\mathbf{W} \in \mathcal{R}^{M\times K}$ is the matrix of weights, with
each row being a prototype.

The learning process is commonly implemented as an
online learning, where each input pattern presentation is followed by an update
of the centroids.
\begin{equation}
    \mathbf{w}_j = \mathbf{w}_j + \eta_i\phi_{i, j} (\mathbf{x}_i - \mathbf{w}_j)
\end{equation}
Since the update does not converge to equilibrium, the SOM algorithm needs
annealing by letting both the learning rate and the neighboring
radius decrease monotonically with epochs (one epoch being a full sequence of
iterations through the dataset) (see the full algorithm in
\ref{alg:som_online}).

\begin{figure}[H]
  \centering
  \includegraphics{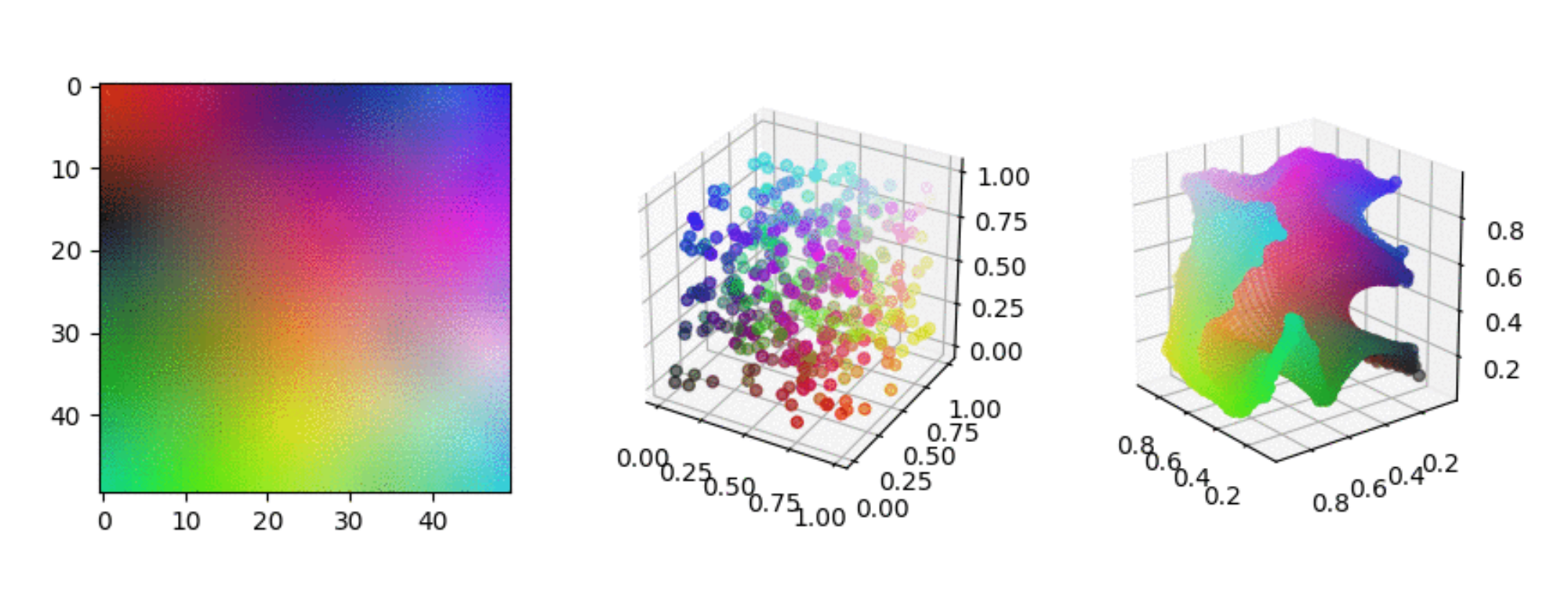}
  \caption{Mapping produced by a SOM with an input layer composed of 3 elements
  (RGB colors), and an inner layer composed of a grid of 50x50 elements. Left:
  the continuous mapping defined by the grid of prototypes (weights of inner
  layer units). Center: the initial dataset, composed of randomly chosen 3D
  points with values in [0,1] on each dimension (each point can be displayed as
  an RGB color). Right: the mapping is produced by finding a 2D manifold in the
  3D space of the data that optimizes the description of all original points.}
  \label{fig:soms}
\end{figure}

SOM allows to build a low dimensional mapping between the input space and the
space of the inner layer (see Figure \ref{fig:soms}). In the described algorithm
implementations (algorithms \ref{alg:som_online} and \ref{alg:kmeans_batch}) a
1-Dimensional space has been chosen for the inner layer. Indeed the radial-basis
function in \ref{eq:som_phi} is based on a distance between two scalar indices.
If for instance a 2-Dimensional description of the hidden space was chosen, the
radial basis function would be based on an euclidean distance between two
2-dimensional points (the current prototype and the winner) in the space where
the inner layer grid was laid out.

The low-dimensional mapping defined by a SOM is discretized, meaning that the
activation of the inner layer can define only discrete points in the space of
the mapping, defined by the position of the units in an imaginary grid that is
laid out on that space. Nevertheless a multivariate interpolation can be applied
(for instance a radial-basis interpolation) so that a pattern of activations in
the inner layer gets related to a single point in the continuous space of the
mapping (see Figure \ref{fig:interpolation}).
\begin{minipage}[c]{\textwidth}
\begin{figure}[H]
\begin{minipage}[c]{.4\textwidth}
\begin{algorithm}[H]
\label{alg:interpolation}
\KwIn{
  $q$: a point in the continuous space of the mapping\newline
}
\KwData{
  $\mathbf{u} \in \mathcal{R}^{K}$: inner layer activations\newline
  $\mathbf{W} \in \mathcal{R}^{K\times M}$: weights\newline
  $\hat{\mathbf{x}} \in \mathcal{R}^{M}$: generated input pattern\newline
  $\sigma$: smoothness
 }
\BlankLine
\Begin{
  \For{$j\gets1$ \KwTo $K$}{
    $u_j \gets \frac{1}{\sigma\sqrt{2\pi}}e^{-\frac{(j - q)^2}{2\sigma^2}}$
  }

  $\hat{\mathbf{x}} \gets \mathbf{W}^T\mathbf{u}$
}
\caption{Radial basis interpolation.}
\end{algorithm}
\end{minipage}%
\begin{minipage}[c]{.6\textwidth}
  \includegraphics{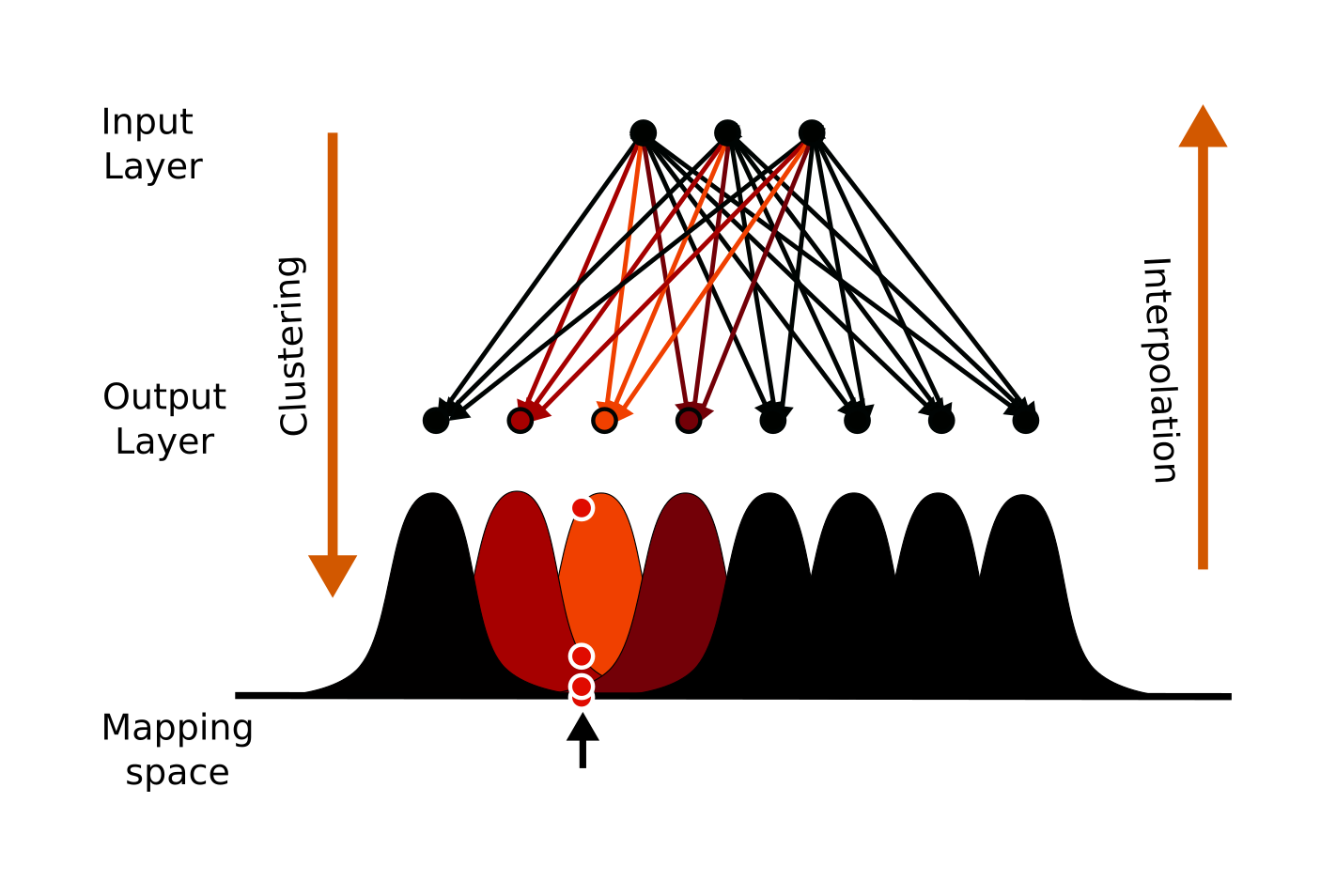}
\end{minipage}
\caption{Application of a radial basis interpolation for the
generation of new data from points in the space of the low-dimensional mapping.
Left: the algorithm for generation. First compute the activation in the inner
layer as radial bases of the given point $q$. Then spread the activations
backward to the input layer. Right: a graphic representation of the
generation-through-interpolation process.}
\label{fig:interpolation}
\end{figure}
\end{minipage}

Comparing the algorithms \ref{alg:kmeans_online} and \ref{alg:som_online} we can
see how they are strictly related. Indeed the SOM algorithm is a minimization of
an energy function very similar to \ref{eq:k-means-energy}, where the $\phi$
operator defined in \ref{eq:k-means-phi} is substituted to the $\phi$ operator
defined in \ref{eq:som_phi} and the weights $\mathbf{w}_j$ stand for the
centroids $\mathbf{c}_j$. Incidentally the SOM learning algorithm can be also be
described with  batch mode (see algorithm \ref{alg:som_batch}).

\begin{minipage}[c]{\textwidth} \RestyleAlgo{ruled}
\begin{algorithm}[H]
\label{alg:som_online}
\KwIn{
  $\mathbf{X} \in \mathcal{R}^{N, M}$: the input dataset\newline
  $\mathbf{W} \in \mathcal{R}^{K, M}$: weights of the SOM (each row is
    a prototype)\newline
  $\eta_{init}$: initial learning rate \newline
  $\sigma_{init}$: initial neighboring radius \newline
  $\tau$: decay window \newline
  $T$: number of epochs
}
\BlankLine
\For{$t\gets0$ \KwTo $T-1$}{
  $\eta\gets\eta_{init}e^{-\frac{t}{\tau}}$\;
  $\sigma\gets\sigma_{init}e^{-\frac{t}{\tau}}$\;
  $\sh{\mathbf{X}}$\;
  \For{$i\gets1$ \KwTo N} {
    $r_{i} = \underset{k}{\argmin} || \mathbf{x}_i - \mathbf{w}_k ||$
    \tcp*[r]{Index of the winner.}
    \For{$j\gets1$ \KwTo $K$} {
      $\phi_{i, j}\gets e^{-\frac{|| j - r_i ||^2}{2\sigma^2}}$
      \tcp*[r]{Update operator.}
      $\mathbf{w}_j\gets\mathbf{w}_j + \eta\phi_{i,j}(\mathbf{x}_i -
        \mathbf{w}_j)$
      \tcp*[r]{Updating of weights.}
    }
  }
}
\caption{SOM learning algorithm - online version}
\end{algorithm}
\end{minipage}

\begin{minipage}[c]{\textwidth} \RestyleAlgo{ruled}
\begin{algorithm}[H]
\label{alg:som_batch}
\KwIn{
  $\mathbf{X} \in \mathcal{R}^{N, M}$: input dataset\newline
  $\mathbf{W} \in \mathcal{R}^{K, M}$: weights of the SOM (each row is
    a prototype)\newline
  $\sigma_{init}$: initial neighboring radius \newline
  $\tau$: decay window \newline
 }
\BlankLine
$t\gets 0$\;
\While{$\sum\limits_{j=0}^K || \Delta \mathbf{w}_j || \ < \epsilon$}{
  $\sigma\gets\sigma_{init}e^{-\frac{t}{\tau}}$\;
  \For{$i\gets1$ \KwTo N} {
    $r_{i} = \underset{k}{\argmin} || \mathbf{x}_i - \mathbf{w}_k ||$
    \tcp*[r]{Index of the winner.}
    \For{$j\gets1$ \KwTo $K$} {
      $\phi_{i, j}\gets e^{-\frac{|| j - r_i ||^2}{2\sigma^2}}$
      \tcp*[r]{Update operator.}
    }
  }

  \For{$j\gets1$ \KwTo $K$} {
    $\mathbf{w}_j \gets \frac{\sum\limits_{i=1}^N\phi_{i,j}\mathbf{x}_i}
    {\sum\limits_{i=1}^N\phi_{i,j}}$
    \tcp*[r]{Updating of clusters.}
  }
  $t\gets t + 1$
}
\caption{SOM learning algorithm -- batch version}
\end{algorithm}
\end{minipage}

\section{Supervised Topological Maps}
\label{sec:stms}
\begin{figure}[ht!]
  \centering
  \includegraphics[width=.6\textwidth]{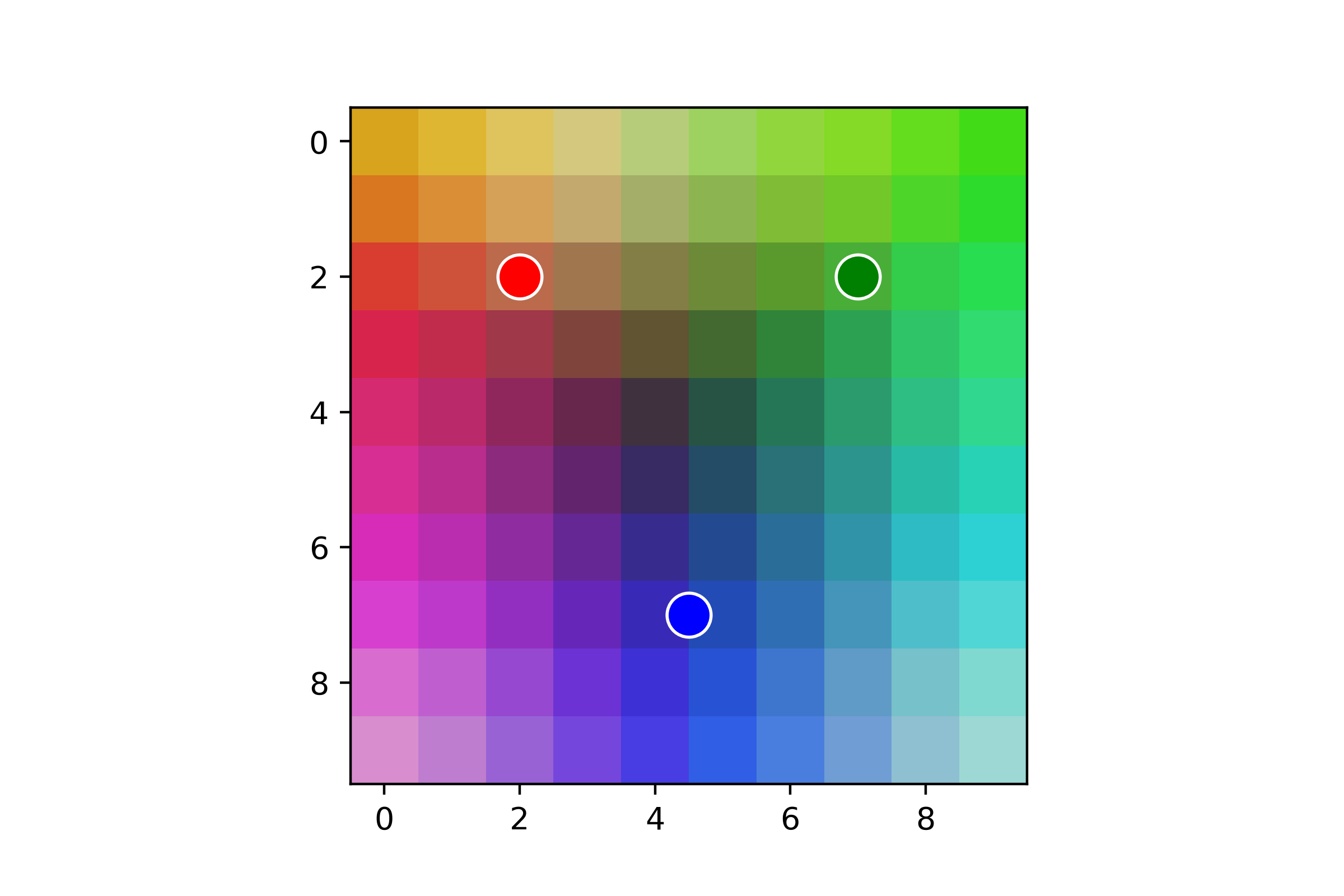}
  \caption{Example of mapping obtained by training an STM with 3 input units
  (RGB) and a 10x10 grid of inner units. On the 2-dimensional mapping on which
  the grid of color prototypes is laid out you can also see the three label
  points categorizing all data-points. Each RGB input pattern is assigned to one
  of the points depending on whether the red, green or blue channel has
  higher amplitude than the others. }
  \label{fig:stm_colors}
\end{figure}

As we saw before, the implementation of the $\phi$ WTA  operator determines
which algorithm between k-means clustering and SOM is used. These two
implementations differ in the smoothness of WTA competition. While in k-means a
simple step function divides the winner prototype from all the others, in SOMs a
smooth function (typically a radial basis) based on the distance of prototypes
from the winner determines the neighborhood of the to-be-updated prototypes. We
will now define another kind of WTA operator depending on 1) the euclidean
distance between input patterns and the prototypes; 2) the euclidean distance
from prototypes and the winner prototype; 3) the euclidean distance of
prototypes from an extrinsically  given point in the low-dimensional mapping
space, for each input pattern:
\begin{align}
  r_i &= \underset{k}{\argmin} || \mathbf{x}_i
-\mathbf{w}_k || \nonumber\\
\label{eq:stm_phi}
\phi_{i, j} &= e^{-\frac{|| j - r_i
||^2}{2\sigma_{r}^2}} e^{-\frac{|| j - t_i ||^2}{2\sigma_{t}^2}}
\end{align}
By means of the WTA operator in eq. \ref{eq:stm_phi}, we can now tell the
learning process where to put the internal representations for a category  of
data. The two parts of the WTA operator will take two different roles in the
update process:1) the radial basis of the distance between the prototypes and
the winner will define the smoothing of prototypes while moving from a label to
another in the mapping space; 2) the radial basis of the distance between the
prototypes and the label points will define the center of attraction for
prototypes of that category. When the update process is complete a low
dimensional mapping emerges in the STM in which prototypes keep close to their
labeling positions depending on their distance from the relative prototype (see
Figure \ref{fig:stm_colors}). Such a mapping can be used to generate new items
starting from any randomly chosen point in the mapping space.

\begin{minipage}[c]{\textwidth}
\RestyleAlgo{ruled}
\begin{algorithm}[H]
\label{alg:stm_batch}
\KwIn{
  $\mathbf{X} \in \mathcal{R}^{N, M}$: input dataset\newline
  $\mathbf{W} \in \mathcal{R}^{K, M}$: weights of the STM (each row is
    a prototype)\newline
  $\sigma_{r_{init}}$: initial neighboring radius \newline
  $\sigma_{t_{init}}$: initial label radius \newline
  $\tau$: decay window \newline
 }
\BlankLine
$t\gets 0$\;
\While{$\sum\limits_{j=0}^K || \Delta \mathbf{w}_j || \ < \epsilon$}{
  $\sigma_r\gets\sigma_{r_{init}}e^{-\frac{t}{\tau}}$\;
  $\sigma_t\gets\sigma_{t_{init}}$\;
  \For{$i\gets1$ \KwTo N} {
    $r_{i} = \underset{k}{\argmin} || \mathbf{x}_i - \mathbf{w}_k ||$
    \tcp*[r]{Index of the winner.}
    \For{$j\gets1$ \KwTo $K$} {
      $\phi_{i, j} = e^{-\frac{|| j - r_i ||^2}{2\sigma_{r}^2}}
        e^{-\frac{|| j - t_i ||^2}{2\sigma_{t}^2}}$
      \tcp*[r]{Update operator.}
    }
  }

  \For{$j\gets1$ \KwTo $K$} {
    $\mathbf{w}_j \gets \frac{\sum\limits_{i=1}^N\phi_{i,j}\mathbf{x}_i}
    {\sum\limits_{i=1}^N\phi_{i,j}}$
    \tcp*[r]{Updating of clusters.}
  }
  $t\gets t + 1$
}
\caption{STM learning algorithm -- batch version}
\end{algorithm}
\end{minipage}

\section{Using STMs}
\label{sec:results}
We give here two examples of the use of STMs for data generation.

\begin{figure}[ht!]
  \centering
  \includegraphics[width=\textwidth]{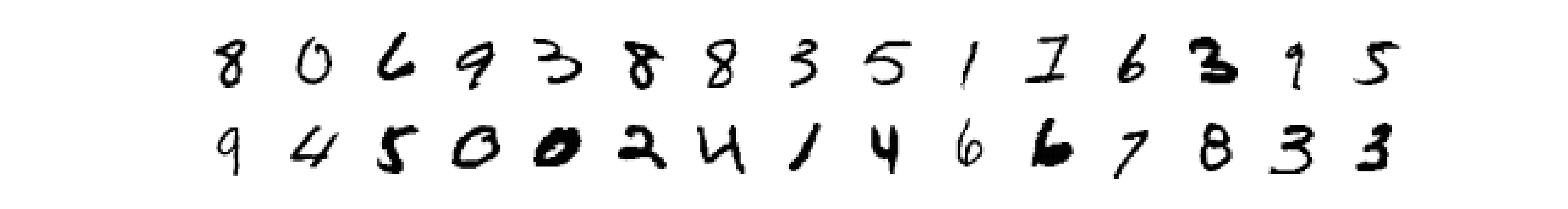}
  \caption{A sample of images from the MNIST dataset of handwritten digits}
  \label{fig:mnist_dataset}
\end{figure}

\begin{figure}[ht!]
  \begin{minipage}[c]{.32\textwidth}
  \centering
  \includegraphics[width=\textwidth]{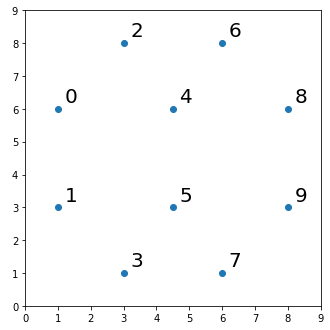}
  a
  \end{minipage}%
  \begin{minipage}[c]{.32\textwidth}
  \centering
  \includegraphics[width=\textwidth]{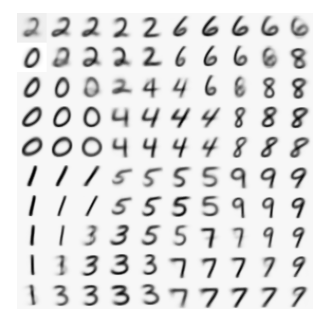}
  b
  \end{minipage}%
  \begin{minipage}[c]{.32\textwidth}
  \centering
  \includegraphics[width=\textwidth]{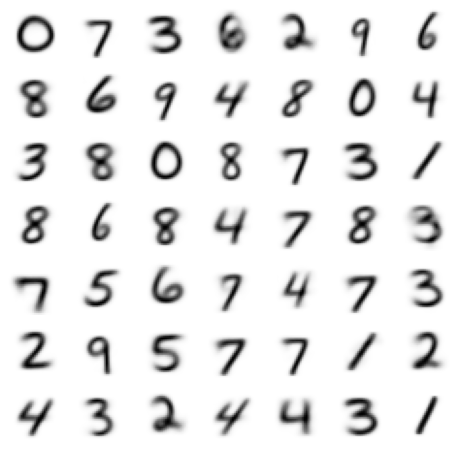}
  c
  \end{minipage}%
  \caption{Mapping produced with the STM algorithm on the MNIST dataset. a) The
  labeling points in the mapping space. Each pattern that is tagged with a digit
  label in the MNIST dataset gets a representation that is close to the
  corresponding point in the mapping. b) Weights of the STM after learning. The
  weights for each inner units are plotted in the corresponding position on the
  inner layer grid; c) images generated from interpolations of a random set
  points in the continuous space of the mapping.}
  \label{fig:stm_mnist}
\end{figure}

First we show how to create a supervised mapping from the MIST dataset of
standard handwritten digits \citep{LeCun1998}. the MNIST dataset is composed of
60.000 28x28 pixel images of the 10 digits (see Figure \ref{fig:mnist_dataset}).
An STM composed of a 28x28 input layer and a 10x10 inner layer was deployed. We
defined the labeling points in the space of the inner layer as in
Figure \ref{fig:stm_mnist}a. After learning the prototypes of the network
 were disposed according to the labeling, as it can be seen in Figure
\ref{fig:stm_mnist}b where the weights of each prototype are plotted in
their position in the grid of the inner layer units. Once the mapping was
obtained we could produce new images through radial-basis interpolation (see
Figure \ref{fig:interpolation}) from randomly chosen points in the
continuous 2-dimensional space of the mapping. Figure \ref{fig:stm_mnist}c
shows some samples of such generated images.

\begin{figure}[ht!]
  \centering
  \includegraphics[width=\textwidth]{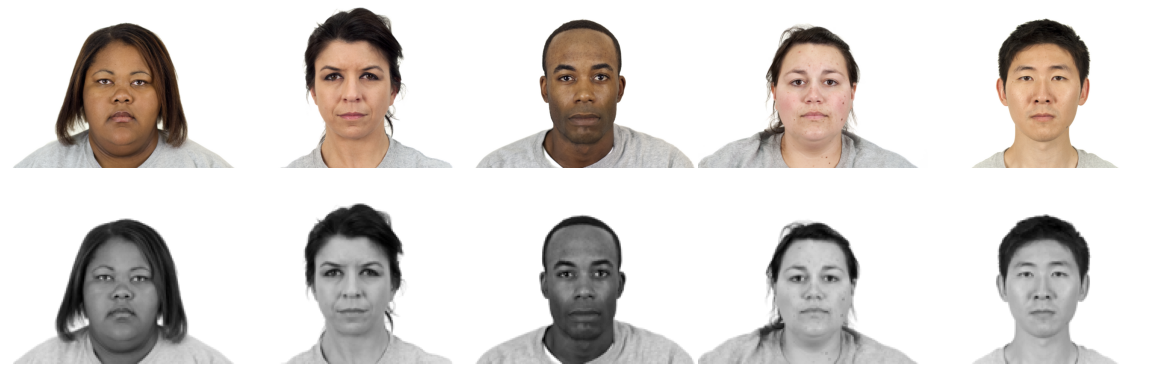}
  \caption{A sample of photographs from the Chicago Face Database. Above, the
  original photographs from the dataset. Below, the processed versions of the CFD
  used in this paper.}
  \label{fig:cdf}
\end{figure}

\begin{figure}[ht!]
  \begin{minipage}[c]{\textwidth}
  \centering
  \includegraphics[width=\textwidth]{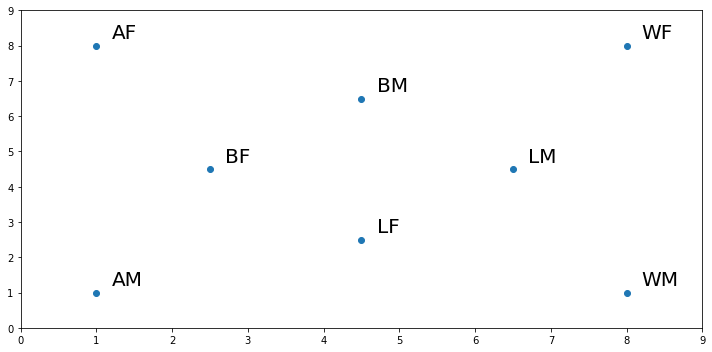}
  a
  \end{minipage}
  \begin{minipage}[c]{\textwidth}
  \centering
  \includegraphics[width=\textwidth]{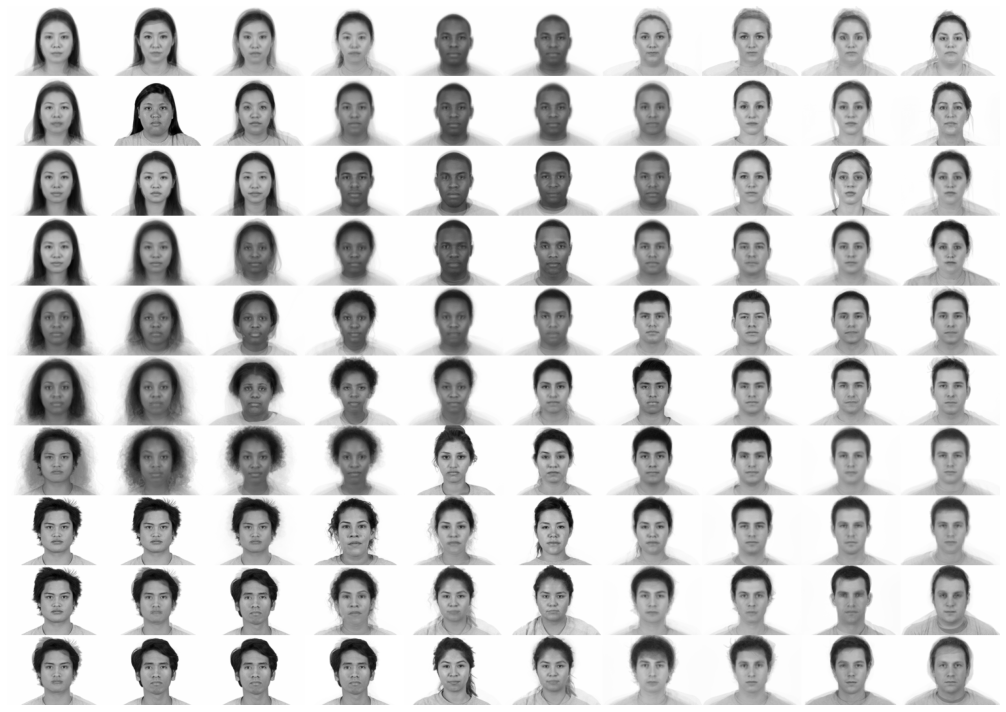}
  b
  \end{minipage}
  \caption{Mapping produced with the STM algorithm on the CFD dataset. a) The
  labeling points in the mapping space. Each pattern that is tagged with a digit
  label in the MNIST dataset gets a representation that is close to the
  corresponding point in the mapping. Labelings are: AF: asian female; AM: asian
  male; BF: black female; BM: black male; LF: latino female; LM: latino male;
  WF: white female; WM: white male. b)  Weights of the STM after learning. The
  weights for each inner units are plotted in the corresponding position on the
  inner layer grid.}
  \label{fig:stm_cfd}
\end{figure}

\begin{figure}[ht!]
  \centering
  \includegraphics[width=\textwidth]{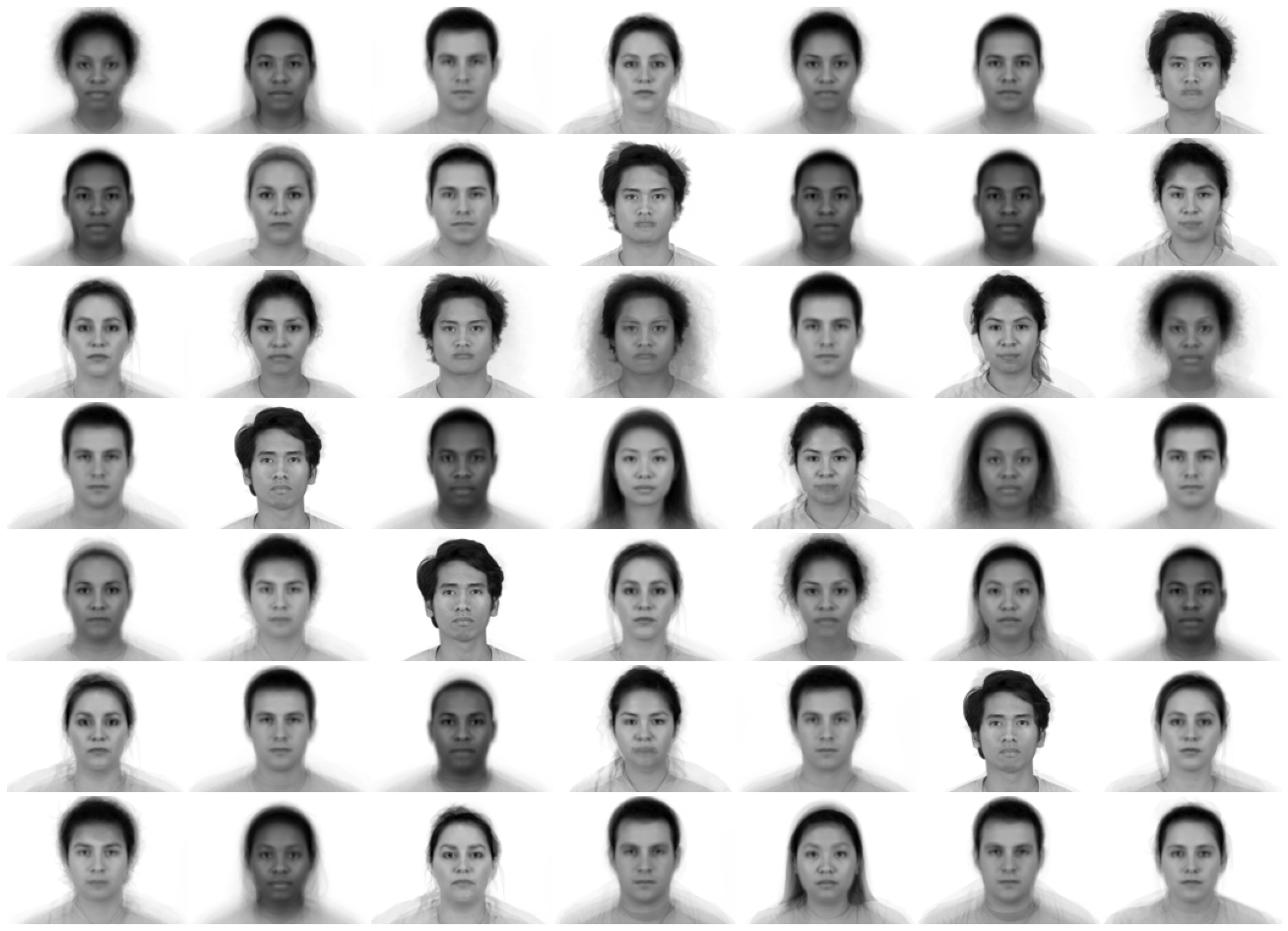}
  \caption{Generation of Images form the CFD mapping. Patterns are generated
  from interpolations of a random set points in the continuous space of the
  mapping.}
  \label{fig:stm_cfd_gen}
\end{figure}

Another example was implemented using the Chicago Face Database \citep[CFD
--][]{Ma2015}. The CFD consists of 158 high-resolution, standardized photographs
of Black and White males and females between the ages of 18 and 40 years . Each
photograph is labeled based on eight categories: asian female; asian male;
black female; black male; latino female; latino male; white female; white male.
The original RGB 2444x1718 pixel photographs were further processed to obtain
152x107 8bit gray-scale images. A sample of the original photographs and
their processed version is shown in Figure \ref{fig:cdf}.
An STM composed of a 152x107 input layer and a 10x10 inner layer was deployed. We
defined the labeling points in the space of the inner layer as in
Figure \ref{fig:stm_cfd}a. After learning the prototypes of the network
 were disposed according to the labeling, as it can be seen in Figure
\ref{fig:stm_cfd}b where the weights of each prototype are plotted in
their position in the grid of the inner layer units. Once the mapping was
obtained we could produce new images through radial-basis interpolation (see
\ref{fig:interpolation}) from randomly chosen points in the
continuous 2-dimensional space of the mapping. Figure \ref{fig:stm_cfd_gen}
shows some samples of such generated images.


\section{Discussion}
\label{sec:discussion}
We showed here a variant of the algorithm used for Self Organizing Maps (SOMs)
which allows to easily obtain a low dimensional mapping of the input space in a
supervised manner, by constraining the positions of the representations in the
mapping to keep themselves close to their corresponding labeling positions. The
new family of algorithms, which we call Supervised Topological Maps (STMs), is
defined by changing the winner-takes-all operator in SOMs so that it is also
sensible to the distance between prototypes and labels in the internal space of
the mapping. The finding that a general family of algorithms can be defined
where the WTA operator identifies the differences between individual algorithms,
have its origins in the strict relation between the k-means algorithm and SOMs.
Indeed, as we showed the two algorithms share many features and the main
difference is the way WTA competition is used. This idea was already implicitly
present in the work of Kohonen, in particular in his description of the Learning
Vector Quantization algorithm \citep[LVQ --][]{Kohonen1995}. LVQ is a
classification algorithm which adapts the position of the winner prototypes
based on their labeling. Following the notation used in this paper we can
describe LVQ by defining its WTA operator as: %
\begin{align}
  \label{eq:lvq-phi}
  r_i &= \underset{k}{\argmin} || \mathbf{x}_i -\mathbf{w}_k || \nonumber\\
  \phi^{\texttt{dist}}_{i,j} & = \left\{
  \begin{array}{ll}
    1 & \texttt{if}\ j = r_i\\
    0 & \texttt{otherwise}
  \end{array}
  \right. \nonumber \\
  \phi^{\texttt{lab}}_{i,j} & = \left\{
  \begin{array}{ll}
    1 & \texttt{if}\ j = t_i\\
   -1 & \texttt{otherwise}
  \end{array}
  \right. \nonumber \\
  \phi^{i, j} & = \phi^{\texttt{dist}}_{i,j} \phi^{\texttt{lab}}_{i,j}
\end{align}
where $i$ is the index of the input pattern, $t_i$ is the label prototype
position  position for the $i^{th}$ input pattern and $j$ is the position of the
prototype currently taken into account. As in SOMs, Although it is a
classification algorithm, LVQ algorithm is strictly related to as k-mean
clustering differing from it in the way the WTA operator is filtered by a
function of the position of prototypes in the cluster space. As in STM its
function of the internal position also depends on an extrinsic labeling of the
cluster space.

We saw how adapting the low-dimensional mapping in a supervised way as in STMs
allows for internal representations of data that are easy to interpret and to be
used for generation of new data. A similar result is nowadays commonly achieved
by using probabilistic generative models such as variational autoencoders
\citep[VAE --][]{Kingma2013} or adversarial autoencoders \citep[AAE
--][]{Makhzani2015}. While these latter algorithms rely on parametric
probability distributions for the shaping of the space of internal
representations, the methods described here are a non-parametric deterministic
way of modeling the space of internal representations. Probabilistic generative
models are a better choice when a robust statistical definition of the
population from which data comes out is required as well a formal level of
reliability that new generated samples come from the same population of given
data. On the other side non-parametric methods as the ones described here seem
to be an easier way to model the internal representations of neural networks for
practical issues. Moreover, the local update methods described here could shed a
new light on the possible ways in which neural mappings from different sensory
or motor modalities are synchronized together in the central nervous systems of
animals.

Another difference from the commonly used neural network algorithms consists in
the fact that STMs as described here are shallow networks and the prototypes
cannot be deeply non-linear functions of the inputs. Nevertheless it could be
shown that a STM layer can substitute the internal layer of a deep neural
network with few changes. It was not the focus of this paper to show such a
possibility. The demonstration of how to use STMs as deep layers a neural
network (e.g. a deep autoencoder) will be the aim of a future work.

Concluding, a new family of algorithms, namely STMs, was described which allows
for  the supervised determination of the internal representations in a neural
mapping.

\section{Acknowledgments}
A special thanks goes to Stefano Zappacosta who reviewed the paper and helped
finding errors in the mathematics. Thanks also go to Gianluca Baldassarre,
Daniele Caligiore, Emilio Cartoni, Vieri santucci and Valerio Sperati for very
helpful discussions and comments on the idea of the paper.

This project has received funding from the European Union's Horizon 2020
Research and Innovation Program under Grant Agreement no. 713010
(GOAL-Robots—Goal-based Open-ended Autonomous Learning Robots). KO, ES, and LJ
were also partially funded by ERC Advanced Grant FEEL, number 323674.

\bibliographystyle{hunsrtnat}
\bibliography{paper}
\end{document}